\documentclass[letterpaper]{article} 
\usepackage{aaai2026}  
\usepackage{times}  
\usepackage{helvet}  
\usepackage{courier}  
\usepackage[hyphens]{url}  
\usepackage{graphicx} 
\usepackage{natbib}  
\usepackage{caption} 
\usepackage{algorithm}
\usepackage{algorithmic}

\usepackage{newfloat}
\usepackage{listings}
\DeclareCaptionStyle{ruled}{labelfont=normalfont,labelsep=colon,strut=off} 
\floatstyle{ruled}
\newfloat{listing}{tb}{lst}{}
\floatname{listing}{Listing}
\title{TEMPLE: Incentivizing Temporal Understanding of Video Large Language Models via Progressive Pre-SFT Alignment}
\author{
    Shicheng Li\textsuperscript{\rm 1}, Lei Li\textsuperscript{\rm 2}, Kun Ouyang\textsuperscript{\rm 1}, Shuhuai Ren\textsuperscript{\rm 1}, Yuanxin Liu\textsuperscript{\rm 1},
    Yuanxing Zhang\textsuperscript{\rm 3},\\
    Fuzheng Zhang\textsuperscript{\rm 4}, Lingpeng Kong\textsuperscript{\rm 2}, Qi Liu\textsuperscript{\rm 2}, Xu Sun\textsuperscript{\rm 1}
}
\affiliations{
    \textsuperscript{\rm 1}National Key Laboratory for Multimedia Information Processing, 
    School of Computer Science, Peking University\\
    \textsuperscript{\rm 2}The University of Hong Kong \\ \textsuperscript{\rm 3}Kling Team \\ \textsuperscript{\rm 4}Kuaishou Technology\\

    \{lisc99,xusun\}@pku.edu.cn
}

\usepackage{booktabs}
\usepackage{multirow}
\newcommand{\METHODNAME}{TEMPLE}

\begin{document}

\maketitle

\begin{abstract}
Video Large Language Models (Video LLMs) have achieved significant success by adopting the paradigm of large-scale pre-training followed by supervised fine-tuning (SFT). 
However, existing approaches struggle with temporal reasoning due to \textbf{weak temporal correspondence in the data} and \textbf{over-reliance on the next-token prediction paradigm}, which collectively result in the absence temporal supervision. 
To address these limitations, we propose \textbf{TEMPLE (TEMporal Preference LEarning)}, a systematic framework that enhances temporal reasoning capabilities through Direct Preference Optimization (DPO). 
To address temporal information scarcity in data, we introduce an automated pipeline for systematically constructing temporality-intensive preference pairs comprising three steps: selecting temporally rich videos, designing video-specific perturbation strategies, and evaluating model responses on clean and perturbed inputs. 
Complementing this data pipeline, we provide additional supervision signals via preference learning and propose a novel Progressive Pre-SFT Alignment strategy featuring two key innovations: a curriculum learning strategy which progressively increases perturbation difficulty to maximize data efficiency; and applying preference optimization before instruction tuning to incentivize fundamental temporal alignment.
Extensive experiments demonstrate that our approach consistently improves Video LLM performance across multiple benchmarks with a relatively small set of self-generated DPO data. 
Our findings highlight TEMPLE as a scalable and efficient complement to SFT-based methods, paving the way for developing reliable Video LLMs.
\end{abstract}

\begin{links}
    \link{Code}{https://github.com/lscpku/TEMPLE}
\end{links}

\section{Introduction}
\label{sec:intro}

Video LLMs have made significant strides in recent years, achieving remarkable success in various video understanding tasks~\cite{gemini,gpt4o}. These advancements follow the standard multi-modal LLM paradigm: First, the model learns visual knowledge and vision-text alignment through large-scale pretraining on image/video-text pairs~\citep{liu2023llava}, and then it acquires instruction following capabilities via supervised fine-tuning (SFT) on curated instruction data~\citep{liu2023llava15,tong2024cambrian,llava-video-sft}. This two-stage approach has enabled impressive improvements in video-language modeling, allowing models to generate coherent, contextually relevant responses conditioned on video input~\cite{videomme,li2023mvbench,timechat}.

Despite this progress, current Video LLMs still exhibit poor temporal understanding, primarily due to two fundamental limitations. First, constructing high-quality video datasets is inherently challenging due to the complex visual and temporal structures in videos. Existing datasets often suffer from weak video-text correspondence, visual shortcuts, or insufficient temporal reasoning tasks~\cite{SingleFrameBias,mangalam2023egoschema,vitatecs}. Second, both pretraining and SFT rely on next-token prediction using ground-truth text, which does not explicitly enforce dynamic temporal understanding. As a result, models frequently neglect subtle yet significant temporal details, over-rely on partial visual or textual cues, and struggle with temporal relationships between events, leading to inconsistent or flawed responses.

\begin{figure*}[t]
    \centering
    \includegraphics[width=0.85\linewidth]{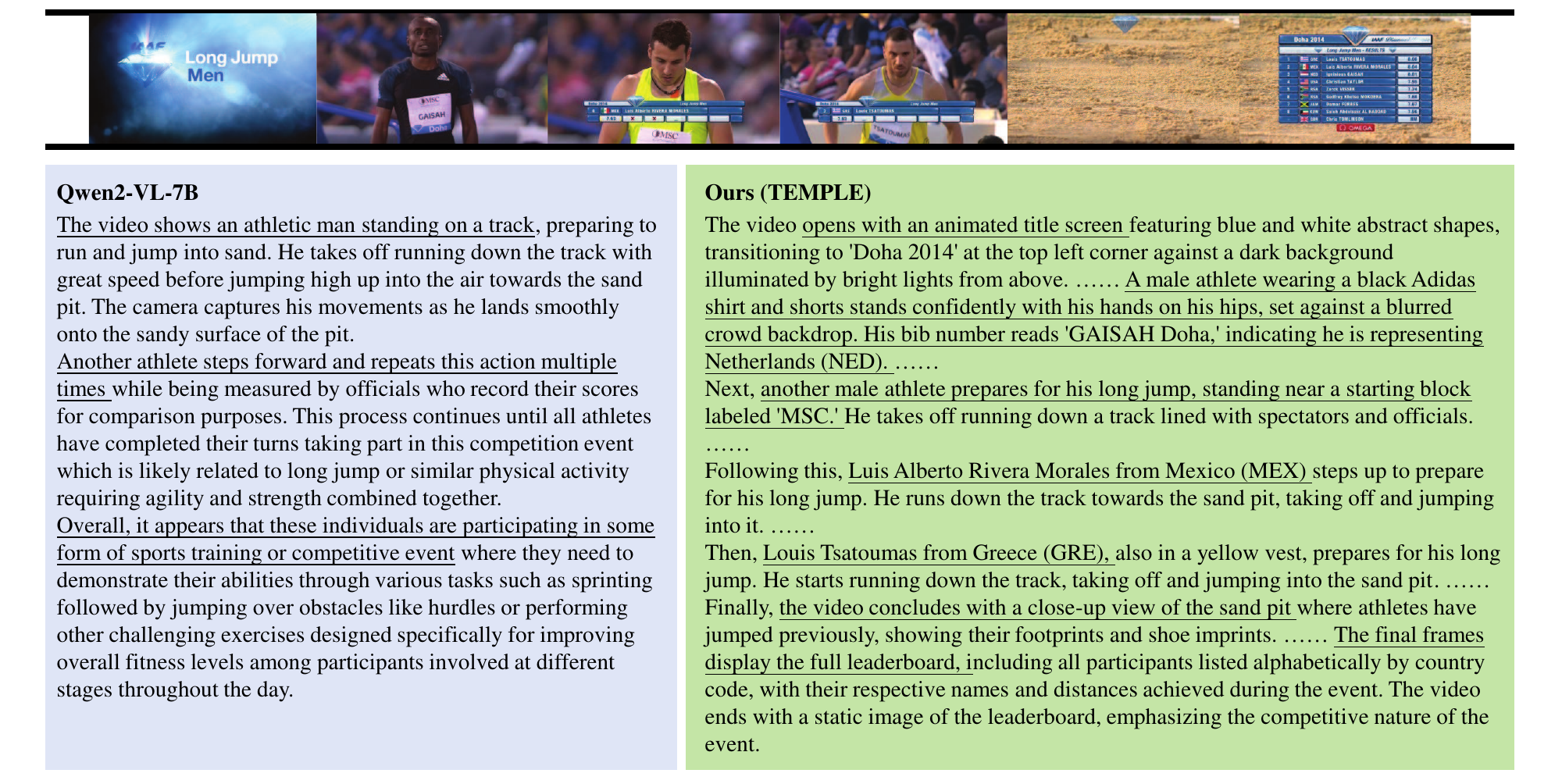}
    \caption{Example of detailed captioning result of the original Qwen2-VL-7B and our approach.}
    \label{fig:case}
\end{figure*}

To address these limitations, we propose \textbf{\METHODNAME{}} (\textbf{TEM}poral \textbf{P}reference \textbf{LE}arning), a systematic framework that tackles both challenges through: (1) a scalable data pipeline that generates high-quality temporal preference data without relying on external LLMs, and (2) a tailored temporal alignment methodology that goes beyond next-token prediction. \METHODNAME{} is built on Direct Preference Optimization (DPO)~\cite{rafailov2023direct}, which optimizes the relative quality of model responses by maximizing the reward margin between preferred and non-preferred outputs. Unlike SFT, which requires high-quality, human-verified responses, DPO focuses on relative comparisons rather than absolute correctness, making it particularly effective for video tasks where perfect annotations are difficult to obtain. 
By learning from temporality-oriented preference pairs, our framework naturally discourages shortcut learning while enabling efficient scaling through automated annotation.

We implement \METHODNAME{} through an automatic preference data pipeline that addresses the scarcity of temporal supervision in video data. 
Inspired by advances in image-based LLMs~\cite{yu2024rlaif,povid}, we generate preference pairs by evaluating model responses under both clean and corrupted visual inputs. Our pipeline consists of three key stages: (1) curating a diverse set of videos that emphasize complex temporal relationships; (2) designing perturbation strategies that precisely target known temporal reasoning weaknesses of current Video LLMs; and (3) constructing preference pairs by contrasting the model's outputs for clean versus temporally corrupted videos. We also introduce a difficulty factor that quantifies perturbation magnitude to control the sample difficulties.
This systematic approach enables us to efficiently generate high-quality contrastive examples that directly highlight temporal understanding capabilities without requiring expensive human annotation.

To leverage the signals in the generated preference data for more effective temporal learning, we further propose a novel Progessive Pre-SFT Alignment strategy tailored for our temporal preference pairs, which consists of two key technical innovations. First, we implement curriculum learning by gradually increasing sample difficulty during preference optimization. This controlled exposure to increasingly challenging temporal contrastive pairs enables more efficient utilization of the preference data. 
Second, we introduce a novel ``Pre-SFT Alignment'' strategy, which reverses the conventional order of model alignment. 
Traditional approaches apply DPO after instruction-tuning~\cite{tunstall2023zephyr}, assuming the model has already learned basic capabilities. In contrast, we argue that robust temporal understanding should be established first, before learning to follow diverse instructions. 
These innovations, coupled with our carefully crafted self-sufficient video filtering and preference generation pipeline, distinguish our work from recent works like Tarsier2~\cite{tarsier2} and TPO~\cite{tpo}, which rely on proprietary LLMs for data filtering and apply DPO in a post-SFT manner.

Through extensive evaluation, we demonstrate that \METHODNAME{} consistently improves model performance across multiple benchmarks including Video-MME~\citep{videomme}, MLVU~\citep{MLVU} and Vinoground~\cite{vinoground} with only a small number of self-generated DPO pairs.
Ablation studies validate the effectiveness of our Progressive Pre-SFT Alignment strategy, which outperforms both the SFT baseline and the conventional SFT-then-DPO approach. 
Further analysis reveals the transferrability of our DPO data across different model architectures and scales, demonstrating the framework's flexibility. 
By introducing this automated, structured preference learning pipeline, we advance the state of Video LLMs and pave the way for more robust and temporally-aware multi-modal models.


\begin{figure*}[t!]
    \centering
    \includegraphics[width=0.75\linewidth]{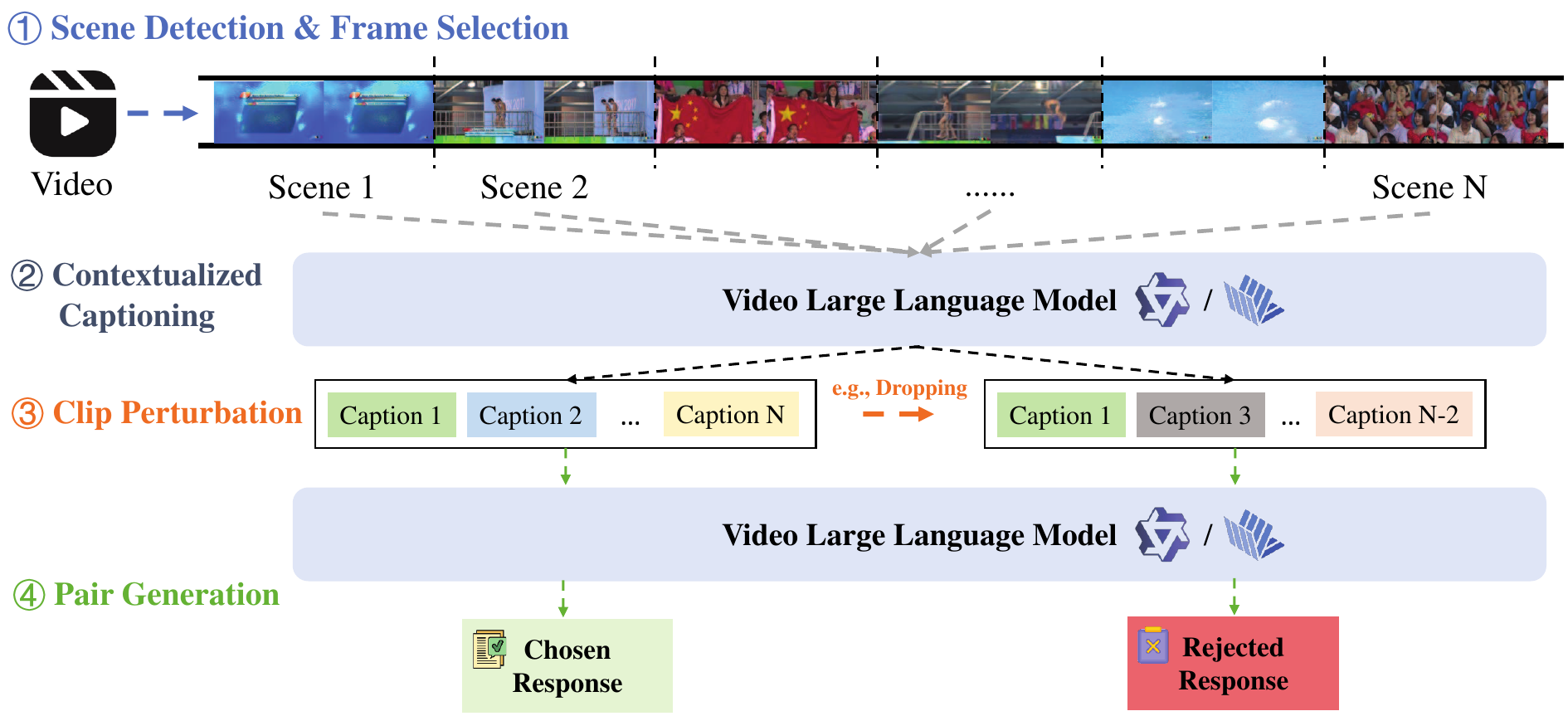}
    \caption{Illustration of our DPO data generation pipeline.}
    \label{fig:workflow}
\end{figure*}

\section{Method}
\label{sec:method}

In this section, we first present a preliminary analysis on the detailed captioning behavior of current Video LLMs, highlighting the need for a more effective alignment strategy. 
We then introduce our DPO-based approach and propose a three-stage pipeline for automatically generating preference data by comparing model responses on clean and perturbed video inputs. 
Finally, we present Progressive Pre-SFT Alignment, where we apply DPO using captioning-based preference pairs with increasing difficulty before SFT, enhancing fine-grained temporal alignment in Video LLMs.

\subsection{Preliminary Analysis on Detailed Captioning}

We conduct a preliminary analysis of the detailed captioning behavior of Qwen2-VL-7B~\cite{Qwen2VL} by manually annotating events in video keyframes and comparing them to events captured in model-generated captions across 19 videos. The comparison reveals a precision of 74.4\% and a recall of merely 35.0\%, indicating that the model not only introduces a considerable amount of hallucinated content but also overlooks a substantial portion of events present in the videos (example shown in Figure \ref{fig:case}). These findings underscore the critical need for improving Video LLMs’ ability to capture and retain temporal information accurately.

\subsection{Preference Data Generation Pipeline}

Video preference data consists of four key components: video, instruction, chosen response, and rejected response. 
Inspired by POVID~\cite{povid}, we generate video DPO data by feeding clean and noised visual inputs into Video LLMs, using the response to the clean input as the chosen response and the response to the corrupted input as the rejected response. To enable preference learning without reliance on external LLMs or APIs, we design a structured pipeline for generating video DPO data, which follows a three-stage process to systematically select videos, craft instructions, and derive response pairs (Figure \ref{fig:workflow}).

\paragraph{Video Selection. }

Public video datasets often contain relatively static videos with minimal variation in visual content, making them easy for models to process but ineffective for learning temporal structure. Such videos provide limited benefit for DPO and are excluded from our selection process. Instead, we prioritize videos featuring multiple, well-separated events with distinct yet connected visual content. To achieve this, we implement a three-step filtering procedure to identify suitable videos and extract clear keyframes.

In the first step, we perform \textbf{scene boundary detection} using TransNetV2~\cite{transnetv2}, which offers more accurate segmentation than tools like PySceneDetect\footnote{\url{https://github.com/Breakthrough/PySceneDetect}}, particularly in handling gradual transitions such as fade-ins and fade-outs.
The number of detected scenes serves as an indicator of a video's temporal complexity, helping us distinguish between dynamic and static content. To refine our selection, we discard monochrome scenes shorter than 0.2 seconds, as these typically correspond to transition frames with little meaningful information. We also filter out videos containing scenes longer than 16 seconds, as such segments often consist of unstructured long shots that are difficult to decompose into discrete events, making them unsuitable for preference learning. Since current segmentation tools struggle to effectively break down these complex videos, we discard them in favor of those with clearer event boundaries.

In the second step, we further filter out videos that contain frequent scene cuts but little new visual information. This includes cases where multiple shots of the same scene from different angles or repetitive back-and-forth cuts, such as those commonly found in interview footage. To achieve this, we introduce a \textbf{similarity grouping} stage, which computes the cosine similarity between middle frames of clips using SigLIP~\cite{siglip} and iteratively group clips with similarity above a predefined threshold. If a video contains too few unique groups, it is deemed to have low information density and is removed. Conversely, if a video contains too many unique groups, it is considered excessively complex, presenting challenges for both computational efficiency and model capability. To balance these factors, we retain only videos with 4 to 32 distinct groups, ensuring sufficient diversity while maintaining manageable complexity. 

In the final step, we \textbf{extract keyframes} that best represent the visual content of each selected clip. Since we have limited clip length to a maximum of 16 seconds, ensuring that each clip corresponds to a single event, we choose to extract two keyframes per clip, which is generally sufficient to capture its core elements. To improve the clarity of the selected keyframes, we set anchor points at one-third and two-thirds of the clip’s duration and select frames that are both close to these anchors and exhibit high visual clarity. Clarity is assessed using the Laplacian operator~\cite{machine_vision}, which measures sharpness to filter out blurry frames. This final step ensures that the extracted keyframes provide distinct and representative snapshots of the video, further enhancing the quality of our DPO data.

Through this three-step process, we systematically filter videos to maximize temporal diversity, eliminate redundant or overly simplistic content, and ensure that selected clips contain clear, well-defined events.

\begin{figure}[t]
    \centering
    \includegraphics[width=\linewidth]{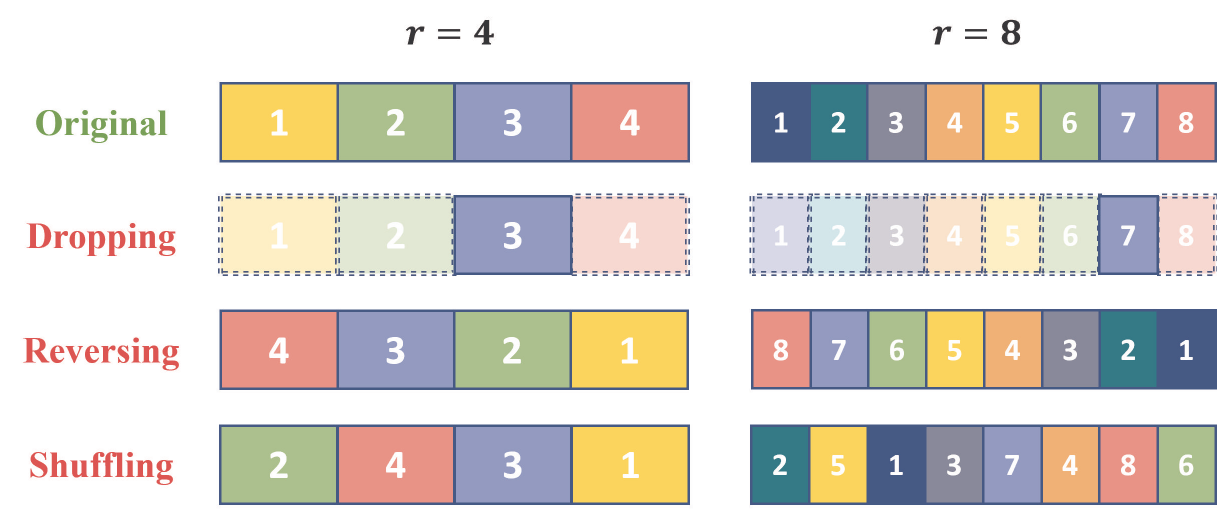}
    \caption{Illustration of video perturbation strategies controlled by the difficulty factor $r$.}
    \label{fig:r}
\end{figure}

\paragraph{Perturbation Strategy and Task Design. }

The design of perturbation strategies and task formulation is critical in our DPO data construction pipeline, as they directly influence the model behaviors we aim to reinforce. Given that video-based QA datasets often suffer from low quality and unreliable annotations, we opt to use detailed captioning as the sole task for preference learning. 

To introduce meaningful preference distinctions, we carefully design perturbation strategies that target the known weaknesses of Video LLMs. One major issue is the model’s tendency to overlook important details in videos or rely on superficial shortcuts in visual or textual inputs. To counter this, we introduce random \textbf{clip dropping}, where certain clips are removed from the video before being fed into the model. This forces the model to generate a response with missing information, leading to a rejected response that either omits critical details or makes unreliable inferences based on partial context. This perturbation explicitly penalizes shortcut reliance and encourages the model to develop a more comprehensive understanding of the input.

Another common failure mode is the difficulty in maintaining temporal structure and distinguishing between visually similar but sequentially distinct frames. To address this, we introduce random \textbf{clip shuffling} and \textbf{clip reversal}, where clips within a video are randomly reordered or flipped temporally. This perturbation preserves the total amount of visual information but disrupts its logical arrangement. Consequently, the model produces a rejected response that misinterprets event chronology or blends together unrelated segments. By optimizing for preference under this perturbation, we encourage the model to develop stronger temporal comprehension and robustness against structural inconsistencies.

These perturbation strategies serve as adversarial challenges, enabling the model to learn more precise and reliable temporal understanding through preference optimization.

\paragraph{Response Generation. }

Generating high-quality, detailed captions for videos with multiple events is a challenging task for Video LLMs, as they often struggle to retain and integrate information across extended sequences. To mitigate this issue, we draw inspiration from prior work in video captioning and adopt a contextual captioning strategy that enables the model to capture fine-grained temporal dependencies without overwhelming its processing capacity.

Instead of requesting a single caption for the entire video, we employ a \textbf{contextualized clip captioning} approach~\cite{chen2024sharegpt4video}. Specifically, at each step, we provide the model with two consecutive clips (corresponding to four frames) and prompt it to generate a caption for the latter while considering the contextual information from both. This localized approach ensures that captions remain detailed while maintaining continuity across the video. 

Once individual clip captions are generated, we aggregate them to form a global video-level caption by feeding all clip captions into the model and prompt it to generate a comprehensive summary of the entire video. This multi-stage process ensures that the final caption preserves both fine-grained event details and high-level narrative coherence.

For response generation in the clean input setting, we directly use the model's output given the full sequence of clip captions. For perturbed inputs, we modify the clip captions according to our designed perturbations.
The model is then prompted with the altered captions to generate a corresponding response, which serves as the rejected response in our preference learning pipeline. By structuring response generation in this manner, we ensure that our preference pairs reflect meaningful variations in video comprehension, reinforcing the model’s ability to process complex temporal relationships and mitigate common failure cases.

\begin{figure}[t]
    \centering
    \includegraphics[width=\linewidth]{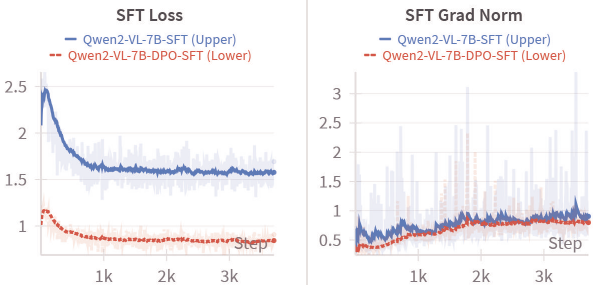}
    \caption{SFT loss and gradient norm. Pre-SFT Alignment leads to lower loss and more stable gradients.}
    \label{fig:sft}
\end{figure}

\subsection{Progressive Pre-SFT Alignment}

\paragraph{Progressive Diffuculty Scheduling. }

To systematically examine the impact of perturbation difficulty in DPO, we introduce a difficulty factor 
$r$ that controls the magnitude of noise applied during preference data generation. This parameter enables us to adjust the degree of disruption in a fine-grained manner, ensuring a structured and progressive learning process for the model.

Given a video with $N$ clips, we regulate difficulty as follows:
(i) For clip dropping, we retain only $\lceil N/r \rceil$ clips while discarding the rest, reducing the available visual information.
(ii) For clip shuffling and reversal, we partition the clips into $\lceil N/r \rceil$ groups, treating each group as an indivisible unit during perturbation. This ensures that lower values of $r$ introduce finer-grained perturbations while preserving more of the original structure, whereas higher values of $r$ cause more drastic disruptions.

Figure \ref{fig:r} illustrates how the videos are perturbed for different difficulty factors.
Intuitively, a smaller $r$ results in weaker perturbations, meaning more information and temporal consistency are preserved, making it harder to differentiate between clean and perturbed inputs. Conversely, a larger $r$ leads to stronger perturbations, making distinctions more apparent and simplifying the preference learning task.

Based on this difficulty factor, we adopt a curriculum learning strategy, where $r$ is gradually decreased throughout training. This allows the model to first learn from easier cases with clearer distinctions and progressively refine its ability to handle more subtle perturbations. By increasing difficulty over time, we ensure that the model develops a more nuanced understanding of video content, leading to improved alignment with fine-grained temporal structures.

\paragraph{Pre-SFT Alignment. }

In the conventional training paradigm, DPO is typically applied after SFT. This ordering ensures that the model first acquires the fundamental ability to perform tasks and follow instructions before refining its alignment with human preferences. However, considering our objective of improving the model’s fine-grained understanding of video content and its temporal structure, we propose an alternative strategy: performing DPO before SFT.
Our key insight is that the primary limitation of Video LLMs lies in their inability to establish precise correspondences between visual input and textual descriptions. Without a strong foundation in fine-grained video-text alignment, the model is prone to relying on static shortcuts and generating responses that are only loosely connected to the true dynamics of the visual content. 
By applying DPO before SFT, we enable the model to first internalize the intricate relationships between video frames and textual descriptions before moving on to more general instruction-following training.
As shown in Figure \ref{fig:sft}, we observe that models trained with Pre-SFT Alignment exhibit significantly lower loss values and more stable gradient norms during the subsequent SFT phase. This suggests that early-stage temporal alignment enhances the model’s ability to process visual information effectively, leading to improved convergence and overall training stability. 
By shifting alignment to an earlier stage in training, our approach ensures that Video LLMs develop a robust foundation in temporal understanding, making them better equipped to handle complex multi-modal reasoning tasks.


\begin{table*}[t]
    \centering
    \small
    \begin{tabular}{l|c|cccc|c}
    \toprule 
        Model & \# Params & VideoMME (TP/TR) & MLVU & Vinoground\_txt & TempCompass\_mc & Average \\
    \midrule 
        Gemini-1.5-pro & - & 75.0 & - & 35.8 & 63.9 & - \\
        GPT-4o & - & 77.2 & 64.6 & 54.0 & 80.0 & 67.7 \\
    \midrule
        VideoLLaMA2 & 72B & 62.4 & 61.2 & 36.2 & - & - \\
        TimeMarker & 7B & 57.3 & 63.9 & - & 65.8 & - \\
        Video-T3 & 7B & 55.0 & 58.1 & 31.6 & 61.2 & 51.5 \\
        LongVA & 7B & 52.6 & 56.3 & 21.2 & 56.1 & 46.6 \\
        LLaMA-3-VILA1.5 & 8B & 45.9 & 22.4 & - & 56.4 & - \\
        LLaVA-NeXT-Video & 34B & 60.2 & - & 23.0 & 68.7 & - \\
        LLaVA-OneVision & 7B & 58.2 & 64.7 & 41.6 & 64.8 & 57.3 \\
        MiniCPM-V-2.6 & 8B & 60.9 & - & 32.6 & 63.0 & - \\
        Video-LLaVA & 7B & 39.9 & 47.3 & 24.8 & 45.6 & 39.4 \\
    \midrule
        Qwen2-VL-2B & 2B & \textbf{55.2} (54.5/34.5) & 57.7 & 26.4 & 64.6 & 51.0 \\
        \quad + SFT & 2B & 54.6 (65.5/32.8) & 60.0 & 27.4 &\textbf{65.0} & 51.8 \\
        \quad + \METHODNAME{} (Ours) & 2B & 55.1 (63.6/35.0) & \textbf{60.2} & \textbf{28.0} & 64.6 & \textbf{52.0} \\
    \midrule
        Qwen2-VL-7B & 7B & 60.4 (70.9/42.4) & 64.5 & 34.0 & 69.6 & 57.1 \\
        \quad + SFT & 7B & 60.7 (70.9/44.1) & 64.6 & 34.6 & 70.8 & 57.7 \\
        \quad + \METHODNAME{} (Ours) & 7B & \textbf{61.0} (74.5/45.2) & \textbf{65.8} & \textbf{37.2} & \textbf{71.4} & \textbf{58.9} \\
    \bottomrule
    \end{tabular}
    \caption{Benchmark accuracy of our approach on VideoMME, MLVU, Vinoground (Text), and TempCompass (Multi-Choice). TP and TR denotes Temporal Perception and Temporal Reasoning, respectively, two subtasks of VideoMME.}
    \label{tab:main_exp}
\end{table*}

\section{Experiment}
\label{sec:experiment}
\subsection{Setting}

We conduct our main experiments using the Qwen2-VL model family~\cite{Qwen2VL}, focusing on two model scales: 2B and 7B parameters, due to its strong performance in open-source Video LLMs. 
We also provide additional evaluation results on Qwen2.5-VL-7B and MiMo-VL-7B in the Appendix. 

\paragraph{Data Generation.} 
The videos used for preference data generation are sourced from a subset of the open-domain LLaVA-Video-178K dataset~\cite{llava-video-sft}, i.e., Youtube videos within the duration of 1-3 minutes. 
After our data pipeline, the final data contains 6,385 preference pairs for each difficulty level, resulting in a total of 25,540 training samples.

\paragraph{Training Details. }
For DPO training, we adopt a four-stage progressive difficulty schedule, where the model is trained sequentially on increasingly challenging data. Beginning with the easiest perturbation setting ($r = 16$), the model gradually adapts to more complex data at $r = 8$, $r = 4$, and finally $r = 2$. 
For SFT, we use a dataset of 60,000 samples randomly drawn from LLaVA-Video-178K~\citep{llava-video-sft}. 
To ensure efficient training, we apply LoRA~\cite{hu2021lora} in both DPO and SFT stages. 
Detailed hyperparameters can be found in the Appendix. 

\paragraph{Evaluation Datasets} For evaluation, we assess models trained with and without DPO on four widely used benchmarks. VideoMME~\cite{videomme} serves as a comprehensive benchmark for general video understanding; MLVU~\cite{MLVU} focuses on long-video comprehension; Vinoground~\cite{vinoground} and TempCompass~\cite{tempcompass} specifically evaluate fine-grained temporal reasoning. For all tasks, we use a maximum pixel per frame of 250,000 and limit the number of frames to 64. 

\paragraph{Baselines} We validate the effectiveness of our method by comparing its performance against two baselines: the original model without additional tuning, and the model trained directly with SFT. We also report the performance of recent Video LLMs as reference points, including a range of open-source models~\cite{videollama2,timemarker,videot3,zhang2024longva,lin2023vila,li2024llava,minicpmv,lin2023video} and closed-source models~\cite{gemini,gpt4o}. 

\subsection{Results}

Table \ref{tab:main_exp} summarizes the results of our main experiments. On the four benchmarks evaluated, our TEMPLE consistently improves video understanding performance compared to the original model and the model trained solely using SFT. 
This demonstrates that performing temporal alignment prior to SFT effectively boosts model performance, yielding greater gains compared to the SFT approach. 
The improvements are particularly pronounced on dimensions related to temporal understanding. On the VideoMME benchmark, TEMPLE delivers substantial gains in temporally-sensitive metrics. For instance, using the Qwen2-VL-7B model, TEMPLE achieves significant improvements over the original model: +3.6 points (70.9 → 74.5) in Temporal Perception and +2.8 points (42.4 → 45.2) in Temporal Reasoning. Furthermore, on the temporally-focused Vinoground benchmark, TEMPLE enhances performance by 3.2 points, representing a relative increase of nearly 10\%. 
These empirical results collectively validate the effectiveness of the TEMPLE framework in enhancing models' temporal alignment capabilities and their overall video comprehension performance, leading to consistent gains across diverse benchmarks, especially in temporal understanding tasks.

\begin{figure}[t]
    \centering
    \includegraphics[width=\linewidth]{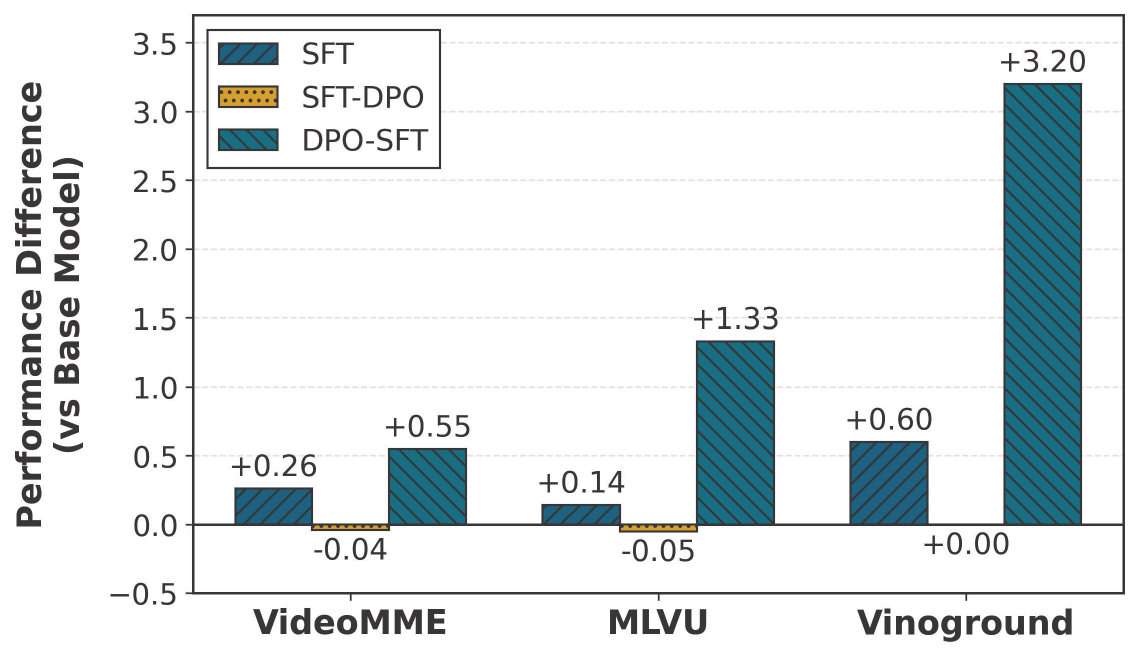}
    \caption{Performance comparison between different tuning strategies on VideoMME, MLVU, and Vinoground (Text). }
    \label{fig:alignment}
\end{figure}

\begin{figure}[t]
    \centering
    \includegraphics[width=\linewidth]{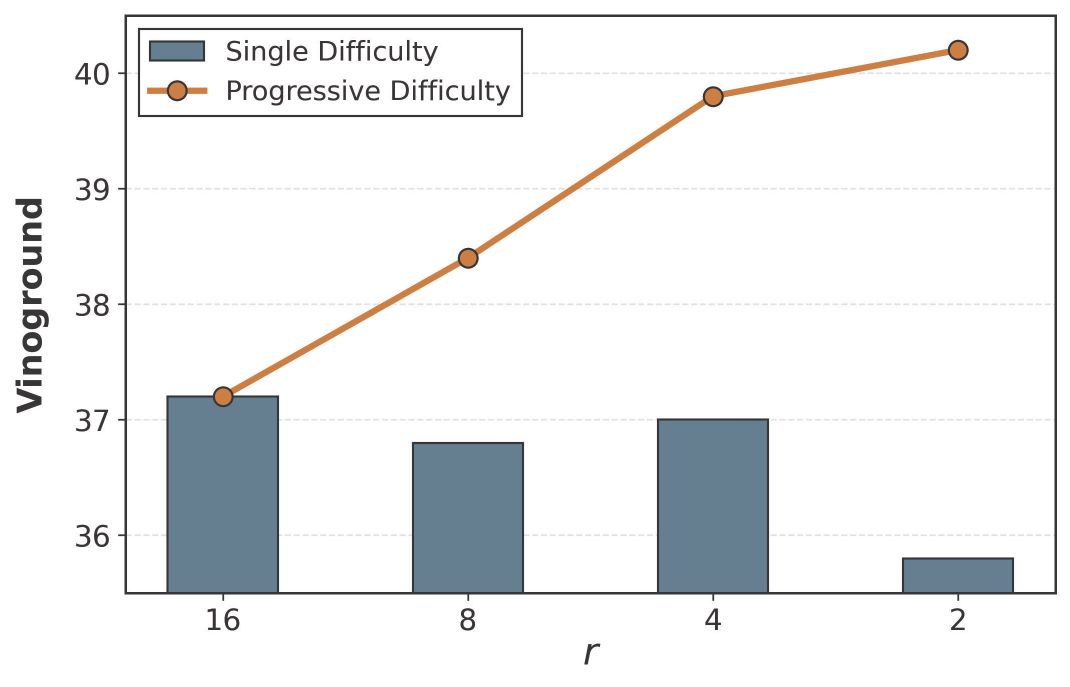}
    \caption{Vinoground (Text) score for models trained with DPO data of different difficulty levels.}
    \label{fig:difficulty}
\end{figure}

\subsection{Analysis}
\paragraph{Ablation Study on Pre-SFT Alignment. }
To showcase the effectiveness of Pre-SFT Alignment, we conduct an ablation study comparing the performance of multiple different tuning orders. Figure \ref{fig:alignment} shows the performance differences of Qwen2-VL-7B, comparing its results after tuning with various strategies against the original base model. The Pre-SFT Alignment strategy consistently outperforms other approaches across all three benchmarks. This demonstrates its ability to effectively combine the benefits of both realms: enhancing the model’s temporal understanding while preserving high levels of instruction-following capability. 
Furthermore, we provide a comparison between our method against previous post-SFT video alignment methods including TPO~\cite{tpo} and LLaVA-Hound~\cite{llava-hound} in the Appendix, which demonstrates the superiority of our method and the complementarity of both strategies. 
This ablation study underscores the advantages of aligning the model with fine-grained temporal knowledge before instruction fine-tuning, enabling better overall performance.

\paragraph{Analysis of Progressive Difficulty Scheduling. }

To investigate the impact of DPO data difficulty on model performance, we first conduct a preliminary comparison by tuning Qwen2-VL-7B with a single split of DPO data containing the same difficulty level and evaluating the models on Vinoground. As shown by the bars in Figure \ref{fig:difficulty}, models tuned with easier DPO data exhibit better performance. This suggests that, due to the initial inability of the model to capture temporal structure, preference optimization with easier data proves more beneficial in the early stages of alignment. This observation motivates our curriculum learning strategy, which formulates DPO as a staged process, progressively increasing the difficulty of DPO data throughout the course. As shown by the line in Figure \ref{fig:difficulty}, this progressive tuning approach leads to steady improvements in model performance, validating the efficacy of our strategy in helping the model gradually refine its temporal alignment.

\begin{table}[t]
    \centering
    \small
    \begin{tabular}{l|c|c|ccc}
    \toprule 
         & Video & ML & \multicolumn{3}{|c}{Vinoground}  \\
    \cmidrule{4-6}
        & MME & VU & Text & Video & Group \\
    \midrule 
        Qwen2.5-VL-7B & 63.1 & 64.0 & 41.6 & \textbf{34.6} & \textbf{17.6} \\
        \quad + SFT & 63.1 & 65.5 & 42.8 & 28.8 & 14.6 \\
        \quad + \METHODNAME{}& \textbf{63.8} & \textbf{67.3} & \textbf{44.2} & 25.8 & 12.4 \\
    \midrule
        InternVL2.5-8B & 64.4 & 67.8 & 42.6 & 23.8 & 10.2 \\
        \quad + SFT & \textbf{64.7} & 67.4 & 42.8 & \textbf{24.4} & 10.2 \\
        \quad + \METHODNAME{} & 64.3 & \textbf{67.9} & \textbf{45.0} & \textbf{24.4} & \textbf{10.4} \\
    \bottomrule
    \end{tabular}
    \caption{Experiment results on the transferrability of our approach across model architectures.}
    \label{tab:model_transfer}
\end{table}

\begin{table}[t]
    \small
    \centering
    \begin{tabular}{l|c|c|ccc}
    \toprule 
         & Video & ML & \multicolumn{3}{|c}{Vinoground}  \\
    \cmidrule{4-6}
        & MME & VU & Text & Video & Group \\
    \midrule 
        Qwen2-VL-7B & 60.4 & 64.5 & 34.0 & 26.6 & 10.6 \\
        \quad + SFT & 60.7 & \textbf{64.6} & 34.6 & 26.6 & 9.6 \\
        \quad + \METHODNAME{} 2B & \textbf{60.9} & 64.3 & \textbf{37.4} & \textbf{28.2} & \textbf{12.4} \\
    \midrule
        Qwen2-VL-2B & \textbf{55.2} & 57.7 & 26.4 & 24.6 & \textbf{6.2} \\
        \quad + SFT & 54.6 & \textbf{60.0} & \textbf{27.4} & \textbf{26.0} & 5.4 \\
        \quad + \METHODNAME{} 7B & 52.9 & 59.5 & 26.0 & 24.6 & \textbf{6.2} \\
    \bottomrule
    \end{tabular}
    \caption{Experiment results on the transferrability of our approach across model sizes.}
    \label{tab:size_transfer}
\end{table}

\paragraph{Transferrability across Model Architectures. }
In our main experiment, we adopt a self-generation approach, using the same model for both generating DPO data and performing DPO. 
To assess the transferrability of our DPO data across different architectures, we conduct an additional experiment where the DPO data generated by Qwen2-VL-7B is used to train two other models: 
Qwen2.5-VL-7B~\cite{Qwen2.5VL} and InternVL2.5-8B~\cite{internvl2.5}.
The results, presented in Table \ref{tab:model_transfer}, show that applying DPO on these models using externally generated data still yields improvements over standard SFT on certain metrics, indicating some level of transferability. However, the gains are neither as consistent nor as pronounced as those observed with self-generated data. This suggests that while DPO data can generalize to some extent, aligning the data generation process with the target model leads to more effective learning.

\paragraph{Transferrability across Model Sizes. }
We further investigate the transferability of our approach across model sizes. Specifically, we evaluate the impact of tuning a larger model using DPO data generated by a smaller model and vice versa. As shown in Table \ref{tab:size_transfer}, when applying the DPO data from Qwen2-VL-2B to fine-tune Qwen2-VL-7B, the model demonstrates improvements on Vinoground, slight gains on VideoMME, but a minor performance drop on MLVU compared to the SFT-only baseline. In contrast, when tuning Qwen2-VL-2B with DPO data generated by Qwen2-VL-7B, the model consistently underperforms relative to the SFT model. These results emphasize the importance of self-generated DPO data, as it better aligns with the model’s specific capabilities and learning patterns. The findings suggest that while some benefits can be retained when transferring data across sizes, DPO is most effective when the data is tailored to the model’s own behavior and representations.

\section{Related Work}
\label{sec:related_work}

\paragraph{Video Large Languge Models} builds upon early efforts to integrate visual understanding into powerful language models~\cite{liu2023llava,liu2023llava15,zhu2023minigpt4}, typically consisting of three key components: a visual encoder, a multimodal projector, and a language model~\cite{video-llm-survey}. 
This design enables the model to perform video understanding by generating textual responses conditioned on the video input. 
Efforts to improve Video LLMs include expanding the scale and improving the quality of video datasets~\cite{llava-video-sft,chen2024sharegpt4video,internvl2.5,DBLP:journals/corr/abs-2506-03569}, and extending the applicability of Video LLMs by enabling them to handle long-duration videos~\cite{zhang2024longva,timechat,ren-etal-2023-testa}, high-resolution content~\cite{Qwen2VL,Qwen2.5VL,minicpmv,li2024llava}, or streaming video inputs~\cite{streamingbench}. 
In contrast, our work addresses a more fundamental challenge in Video LLMs: the acquisition of temporal knowledge. Prior studies have frequently highlighted the deficiency of these models in capturing fine-grained temporal structure, which we directly tackle through our proposed approach.

\paragraph{Direct Preference Optimization}
was originally introduced in the text domain to align language model behavior with human preferences. By optimizing an implicit reward model derived from preference data, DPO provides an effective approach to post-training alignment. 
More recently, research~\cite{povid,MIA-DPO,li-etal-2024-vlfeedback,vlrlhf} has adapted DPO and other alignment strategies to multi-modal LLMs by developing carefully curated data generation pipelines. 
In the video domain, some works~\cite{llava-hound,tpo,tarsier2} have also explored DPO for Video LLMs based on model responses to corrupt visual inputs. 
However, our approach distinguishes itself through several key improvements. 
First, our method introduces a carefully designed pipeline that automatically filters temporally rich videos and generates preference data in a fully self-sufficient manner, without relying on external proprietary models. 
Furthermore, beyond dataset curation, we present a novel Progressive Pre-SFT Alignment strategy for more effective temporal alignment, setting our approach apart from concurrent work in video DPO.

\section{Conclusion}
\label{sec:conclusion}

In this work, we propose a novel framework to enhance the temporal alignment of Video LLMs through DPO, addressing the lack of supervision signals in current training paradigms. 
Our method leverages a carefully designed data generation pipeline to curate temporality-oriented video preference data and a novel Progressive Pre-SFT Alignment strategy to inject temporal understanding abilities into Video LLMs. We demonstrate the effectiveness of our approach through extensive experiments on multiple benchmarks, showing substantial gains in temporal understanding. 
These results open the door for further improvements in Video LLMs, particularly in refining their ability to understand and reason about temporal dynamics in videos.

\appendix

\begin{table}[t!]
    \centering
    \small 
    \begin{tabular}{l|ccc}
    \toprule 
         & Original & After Step 1 & After Step 2  \\
    \midrule 
        YouTube 1-2m & 22427 & 8724 & 159 \\
        YouTube 2-3m & 24685 & 2191 & 597 \\
    \midrule
        Total & 47112 & 10915 & 2195 \\
    \bottomrule
    \end{tabular}
    \caption{The number of remaining videos after each step of our video selection pipeline.}
    \label{tab:stats}
\end{table}

\begin{table*}[t!]
    \centering
    \small 
    \begin{tabular}{l|c|c|c|c|c|ccc|cccc}
    \toprule 
        \multirow{2}{*}{Model} & \multirow{2}{*}{VM} & \multirow{2}{*}{ML} & \multirow{2}{*}{LVB} & \multirow{2}{*}{ES} & \multirow{2}{*}{PT} & \multicolumn{3}{c|}{Vinoground} & \multicolumn{4}{c}{TempCompass} \\
        \cmidrule(lr){7-9} \cmidrule(lr){10-13}
        & & & & & & V & T & G & CM & C & MC & YN \\
    \midrule 
    MiMo-VL-7B+SFT       & \textbf{63.52} & \textbf{65.38} & 58.27 & 57.80 & 65.71 & 41.2 & 29.2 & 14.8 & \textbf{82.43} & 54.14 & \textbf{73.54} & 73.09 \\
    MiMo-VL-7B+TEMPLE    & 63.26 & 65.30 & \textbf{58.56} & \textbf{58.60} & \textbf{65.81} & \textbf{43.6} & \textbf{30.0} & \textbf{16.8} & 82.30 & \textbf{60.08} & 73.29 & \textbf{73.18} \\
    \midrule
    Qwen2.5-VL-7B+SFT     & 63.15 & 66.43 & 57.89 & \textbf{55.40} & \textbf{69.50} & 43.8 & 27.2 & 12.6 & 81.84 & 53.69 & \textbf{73.67} & \textbf{74.56} \\
    Qwen2.5-VL-7B+TEMPLE  & \textbf{63.48} & \textbf{67.11} & \textbf{58.56} & 53.80 & 69.01 & \textbf{44.0} & \textbf{29.2} & \textbf{15.0} & \textbf{82.10} & \textbf{55.34} & 73.29 & \textbf{74.56} \\
    \bottomrule
    \end{tabular}
    \caption{Performance of TEMPLE using Qwen2.5-VL-7B-Instruct and MiMo-VL-7B-SFT as base models. }
    \label{tab:add_model}
\end{table*}

\begin{table*}[t!]
    \centering
    \begin{tabular}{l|ccccc|c}
    \toprule 
        Model & VideoMME & MLVU & LongVideoBench & EgoSchema & PerceptionTest & Average \\
    \midrule 
        SFT                 & 60.67 & 64.60 & 55.12 & 50.00 & 60.09 & 58.10 \\
        TEMPLE-SFT          & 60.96 & 65.79 & 55.35 & 52.60 & 60.82 & 59.10 (+1.00) \\
        SFT-TPO             & 60.19 & 63.93 & 56.62 & 52.60 & 60.78 & 58.82 (+0.72) \\
        SFT-Hound           & 60.30 & 63.92 & 55.80 & 51.40 & 60.56 & 58.40 (+0.30) \\
        TEMPLE-SFT-TPO      & 60.70 & 64.63 & 56.40 & 54.80 & 61.49 & 59.60 (+1.50) \\
        TEMPLE-SFT-Hound    & 60.89 & 64.54 & 56.02 & 53.40 & 61.10 & 59.19 (+1.09) \\
    \bottomrule
    \end{tabular}
    \caption{Performance of different combinations of TEMPLE and post-SFT DPO methods on general video understanding benchmarks. }
    \label{tab:dpo-comparison-general}
\end{table*}

\begin{table*}[t!]
    \centering
    \begin{tabular}{l|ccc|cccc|c}
    \toprule 
        \multirow{2}{*}{Model} & \multicolumn{3}{c|}{Vinoground} & \multicolumn{4}{c|}{TempCompass} & \multirow{2}{*}{Average} \\
        \cmidrule(lr){2-4} \cmidrule(lr){5-8}
         & Video & Text & Group & Matching & Captioning & Multi-Choice & Yes/No & \\
    \midrule 
        SFT                 & 34.6 & 26.6 & 9.6  & 80.51 & 46.71 & 70.82 & 74.84 & 45.91 \\
        TEMPLE-SFT          & 37.2 & 29.6 & 12.2 & 79.51 & 47.65 & 71.39 & 74.60 & 47.31 (+1.40) \\
        SFT-TPO             & 34.6 & 25.4 & 9.8  & 80.31 & 49.80 & 70.76 & 74.89 & 46.11 (+0.20) \\
        SFT-Hound           & 33.6 & 28.4 & 9.8  & 80.31 & 49.20 & 70.82 & 74.68 & 46.34 (+0.42) \\
        TEMPLE-SFT-TPO      & 35.6 & 27.2 & 11.2 & 79.51 & 51.50 & 71.33 & 74.40 & 46.93 (+1.02) \\
        TEMPLE-SFT-Hound    & 36.4 & 29.0 & 11.0 & 79.64 & 50.35 & 71.01 & 74.89 & 47.22 (+1.31) \\
    \bottomrule
    \end{tabular}
    \caption{Performance of different combinations of TEMPLE and post-SFT DPO methods on temporal understanding benchmarks. }
    \label{tab:dpo-comparison-temporal}
\end{table*}

\section{Data}

For our experiments, we generate preference data using videos from LLaVA-Video-180K, a comprehensive conversation-style dataset designed for fine-tuning Video LLMs. Its broad coverage of video content and duration makes it an ideal source for our pipeline. However, we observe that a significant portion of videos in this dataset exhibit low dynamic content and insufficient temporal information, a common limitation in video datasets. Examples include static interview footage with alternating speaker shots or multi-view recordings of a single object/action. 

This observation motivates our video filtering pipeline which selects temporally informative videos based on scene number and inter-scene similarity. 
Table \ref{tab:stats} presents the number of remaining videos after each step of our video selection pipeline. The large number of discarded videos, especially for shorter video within the duration of 1-2 minutes, validates our observation and underscores our pipeline’s necessity in generating temporality-oriented preference data.

\section{Training Details}

Our DPO process is split into four stages where each stage corresponds to a single difficulty factor. During DPO, we apply LoRA~\cite{hu2021lora} with rank 32, $\alpha = 64$, a batch size of 16, and a learning rate of 5e-6, while keeping the ViT module frozen. We adopt Adam~\cite{adam} as the optimizer throughout training. Each DPO stage lasts for approximately one hour on 8 NVIDIA H800 GPUs, accumulating to a total of four hours for the entire alignment process.

During SFT, the ViT module is unfrozen to allow for additional adaptation. The SFT phase follows a similar training setup with LoRA rank set to 32, $\alpha$ set to 32, a batch size of 16, and a higher learning rate of 1e-4. The entire fine-tuning process takes approximately three hours on the same hardware setup. 

Both DPO and SFT follow the original Qwen2-VL preprocessing pipeline, where videos are sampled at 2 FPS. Note that while extracted keyframes are used for data construction, they are not used during training. To accommodate memory constraints, we limit the number of pixels per frame to 90,000, the maximum number of frames per video to 100, and the total sequence length to 8,192.

\section{Additional Results}

\paragraph{Evaluation on additional base models. } 
To demonstrate the broad applicability of our method, we apply TEMPLE to Qwen2.5-VL-7B-Instruct and MiMo-VL-7B-SFT and evaluate model performance on an extended set of benchmarks including VideoMME (VM), MLVU (ML), LongVideoBench (LVB), EgoSchema (ES), PerceptionTest (PT), Vinoground (Text, Video, Group), and TempCompass (Caption Matching, Captioning, Multi-Choice, Yes/No)~\cite{videomme,MLVU,LongVideoBench,mangalam2023egoschema,perceptiontest,vinoground,tempcompass}. The results are presented in Table \ref{tab:add_model}. 
Results show that TEMPLE consistently enhances performance over SFT across a majority of the evaluated benchmarks. Notably, for the Vinoground benchmark, TEMPLE improves all three metrics for both base models, highlighting its strength in temporal alignment. Beyond this, TEMPLE also demonstrates gains in general video understanding benchmarks. These improvements, observed across two distinct base models, underscore the generalizability and effectiveness of TEMPLE for enhancing video-language models.

\paragraph{Comparison with other video DPO methods. }
We further compare our Pre-SFT Alignment strategy with other video DPO methods which follow the conventional post-SFT DPO strategy. Specifically, we perform TPO~\cite{tpo} and LLaVA-Hound-DPO~\cite{llava-hound} after training the model with SFT. 
During this post-training process, we adopt the same hyper-parameters as our DPO process in TEMPLE. 
We also experimented with combining TEMPLE with post-SFT DPO methods to see how these two strategies work together. 
The results on general video understanding and temporal understanding benchmarks are presented in Table \ref{tab:dpo-comparison-general} and Table \ref{tab:dpo-comparison-temporal}, respectively. 

These results lead to several key observations. First, all evaluated methods, including TEMPLE, TPO, and LLaVA-Hound-DPO, consistently outperform the SFT baseline across both benchmark categories. 
Second, TEMPLE alone surpasses both TPO and LLaVA-Hound-DPO when applied individually, achieving higher average scores on general video understanding (59.10 vs. 58.82 for TPO and 58.40 for Hound) and temporal understanding benchmarks (47.31 vs. 46.11 and 46.34, respectively), underscoring the advantage of pre-SFT temporal alignment. Third, the two approaches exhibit complementary strengths: TPO and Hound excel in specific tasks such as LongVideoBench and TempCompass, while TEMPLE demonstrates superior performance on others like Vinoground. The orthogonality of these strategies is affirmed by the additional gains observed when combining them, which promotes the joint application of both paradigms for comprehensive performance improvement.

\section*{Acknowledgments}
This research was partially supported by the National Natural Science Foundation of China under Grant No. 92470205 and No. 62176002. Xu Sun is the corresponding author. 

\small
\bibliography{aaai2026}

\end{document}